\documentclass[journal,twoside,web]{ieeecolor}
\usepackage{lcsys}
\usepackage{cite}
\usepackage{amsmath,amssymb,amsfonts}
\usepackage{algorithmic}
\usepackage{graphicx}
\usepackage{textcomp}
\def\BibTeX{{\rm B\kern-.05em{\sc i\kern-.025em b}\kern-.08em
    T\kern-.1667em\lower.7ex\hbox{E}\kern-.125emX}}
\usepackage{algorithm}
\usepackage{algorithmic}

\usepackage{amsmath}
\usepackage{mathtools}
\usepackage{amsfonts}
\usepackage{amssymb}
\usepackage{color}
\usepackage{svg}

\usepackage{graphicx}

\usepackage{caption}
\let\labelindent\relax
\usepackage{enumitem}
\usepackage{balance}

\DeclareMathOperator*{\argmin}{arg\,min}

\makeatother

\newtheorem{definition}{Definition}
\newtheorem{lemma}{Lemma}
\newtheorem{remark}{\bf Remark}
\newtheorem{assumption}{Assumption}
\newtheorem{theorem}{Theorem}

\DeclareGraphicsExtensions{.png,.jpg,.eps}

\newcommand{\xddots}{%
	\raise 4pt \hbox {.}
	\mkern 10mu
	\raise 1pt \hbox {.}
	\mkern 10mu
	\raise -2pt \hbox {.}
}

\begin{document}

\title{Learning Dissipative Neural Dynamical Systems
}

\author{Yuezhu Xu and S. Sivaranjani$^{*}$  
\thanks{*Corresponding author.}
\thanks{The authors are with the School of Industrial Engineering, Purdue University, West Lafayette, IN 47907, USA.
      (emails: {\tt\small xu1732@purdue.edu}, {\tt\small sseetha@purdue.edu}). 
       } %
       \thanks{This work was partially supported by the Air Force Office of Scientific Research grant, FA9550-23-1-0492.}
}

\pagestyle{empty}
\maketitle
\thispagestyle{empty}
\begin{abstract}

Consider an unknown nonlinear dynamical system that is known to be dissipative. The objective of this paper is to learn a neural dynamical model that approximates this system, while preserving the  dissipativity property in the model. In general, imposing dissipativity constraints during neural network training is a hard problem for which no known techniques exist. In this work, we address the problem of learning a dissipative neural dynamical system model in two stages. First, we learn an unconstrained neural dynamical model that closely approximates the system dynamics. Next, we derive sufficient conditions to perturb the weights of the neural dynamical model to ensure dissipativity, followed by perturbation of the biases to retain the fit of the model to the trajectories of the nonlinear system. We show that these two perturbation problems can be solved independently to obtain a  neural dynamical model that is guaranteed to be dissipative while closely approximating the nonlinear system. 
\end{abstract}


\section{Introduction}
\IEEEPARstart{T}{he} identification of dynamical system models for control, in both linear and nonlinear settings, is a long-studied problem \cite{gevers2005identification}. Typically, nonlinear systems have been modeled using approximate linear models \cite{ljung2001estimating} or linear parameter varying models \cite{dos2011linear}, and more recently, as high-dimensional linear approximations using Koopman operator models \cite{mauroy2016linear,korda2018linear} for the purposes of analysis and control design. Deep learning-based dynamical system models, such as neural ordinary differential equations (neural ODEs) \cite{chen2018neural} and physics-informed neural networks \cite{wang2021physics} to capture the dynamical behavior of nonlinear systems have also recently gained attention. 

When identifying models for control, it is typically not sufficient to simply obtain a model that approximates the system's dynamical behavior. Rather, we would ideally like to preserve essential system properties such as stability in the identified models. One such control-relevant  property that is particularly useful is dissipativity \cite{willems1972dissipative,willems1972dissipative2}. Dissipativity is an input-output property that generalizes the notion of energy storage in circuit theory to a general quadratic form that is applicable to any nonlinear system. Dissipativity
is an attractive property because dissipative systems are stable and generally robust to  perturbations \cite{hill1976stability,hill1980dissipative}. Further, dissipativity provides a general framework to guarantee several crucial properties like $\mathcal{L}_2$ stability, passivity, conicity, and sector-boundedness. Recently, dissipativity theory has also been extended to analyze safety and reachability \cite{coogan2014dissipativity,arcak2016networks}. Finally, dissipativity is a compositional property that is preserved for certain interconnection topologies, and has been widely exploited for scalable, distributed, and compositional control synthesis in networked systems \cite{antsaklis2013control}-\nocite{agarwal2020distributed,agarwal2019sequential}\cite{arcak2022compositional}  and switched/hybrid/event-triggered systems \cite{lavaei2022dissipativity}-\nocite{agarwal2016dissipativity,awan2019dissipativity}\cite{liu2019event}, and has found applications in several domains, including but not limited to, electromechanical systems \cite{ortega2013passivity} robotics \cite{hatanaka2015passivity}, power grids \cite{sivaranjani2020distributed,sivaranjani2020mixed}, and process control \cite{bao2007process}. Therefore, we would ideally like to preserve this control-relevant property during system identification.

Here, we consider the problem of learning a neural dynamical system model for an unknown nonlinear system that is known a priori to be dissipative. We focus on neural dynamical systems for the following reason. Neural networks (NNs) are universal function approximators \cite{hornik1989multilayer}; therefore, neural dynamical models can capture nonlinear dynamical behavior well beyond the `local' region in the vicinity of the equilibrium that is captured by linear models, allowing us to expand the validity and usefulness of our control designs. However, there are limited guarantees on control-relevant properties such as stability, robustness, or dissipativity that can be obtained using such learning-based models.

System identification, even in linear settings, does not automatically preserve properties like dissipativity and passivity without explicit constraints, even if the original system is known a priori to possess such properties. While identification of stable models has been studied for several decades, system identification approaches that preserve system dissipativity and passivity properties have only been investigated in the context of linear systems (see \cite{grivet2015passive} for a comprehensive survey),  linear approximations for nonlinear systems \cite{sivaranjani2022data,sivaranjani2020data}, and Koopman operator models \cite{hara2020learning,khosravi2023kernel}. Learning stable neural ordinary differential equation (ODE) models has been achieved through neural Lyapunov functions or Lyapunov constraints (see \cite{nghiem2023physics} for a compilation of works addressing this topic). There is also some recent work on learning dissipative neural dynamics limited to specific port-Hamiltonian network structures; further, these models only apply when the system inputs are constant \cite{zhong2020dissipative}. Dissipativity theory for neural dynamical systems has also been confined to special cases such as Lyapunov stability  for autonomous systems, that is, systems without inputs. \cite{drgovna2022dissipative}. The problem of \textit{learning provably dissipative deep neural dynamical  models} for general nonlinear systems, especially in the closed-loop setting, remains an open problem. The key challenge lies in imposing matrix inequality constraints, such as those required to guarantee dissipativity, during deep NN training; this is a hard problem with no known solution. 

In this work, we address the particular problem of learning a dissipative neural dynamical  model for a nonlinear system that is known to satisfy an incremental dissipativity property. We propose a two-stage solution to address this problem. First, we train an unconstrained feedforward deep neural ODE model using input-output trajectories from the nonlinear system. Next, we derive sufficient conditions on the weights of the NN to guarantee incremental dissipativity of the learned model, and pose an optimization problem to minimally perturb the weights to enforce these conditions. Finally, we adjust the biases, as necessary, to retain the fit of the dissipative neural dynamical model to the ground truth. 

The key contributions of this work are as follows. First, we derive sufficient conditions to guarantee incremental dissipativity of deep neural dynamical models. 
Second, we propose an algorithm where dissipativity can be imposed by perturbation of the weights alone, allowing us to independently tune the biases to retain the fit of the model to the true system dynamics without losing our dissipativity guarantee. To the best of our knowledge, this is the first work to develop algorithms that preserve  the \textbf{input-output property} of dissipativity in identification of neural ODEs (where the theory has been limited to autonomous systems).

This paper is organized as follows. We formulate the dissipative neural ODE identification problem 
in Section \ref{sec: problem}, and present a two-stage solution in Section \ref{sec: learning}. We present a case study demonstrating the approach in Section \ref{sec: case}, and discuss directions for future work in Section \ref{sec: future}. The proofs of all results are presented in the Appendix.

\textit{Notation:} We denote the sets of real numbers, positive real numbers including zero, and $n$-dimensional real vectors by $\mathbb{R}$, $\mathbb{R}_{+}$ and $\mathbb{R}^{n}$ respectively. Define $\mathbb{Z}_N=\{1,\ldots,N\}$, where $N$ is a natural number excluding zero. Given a matrix $A \in \mathbb{R}^{m \times n}$,  $A^T\in \mathbb{R}^{n \times m}$ represents its transpose. A symmetric positive definite matrix $P \in \mathbb{R}^{n \times n}$ is represented as $P>0$ (and as $P\geq 0$, if it is positive semi-definite). Similarly, a symmetric negative definite matrix $P \in \mathbb{R}^{n \times n}$ is represented as $P<0$ (and as $P\leq 0$, if it is negative semi-definite). The standard identity matrix is denoted by $\mathbf{I}$, with dimensions clear from the context. Given two vectors $x,y\in \mathbb{R}^n$, we define the operator $\delta(x,y)=y-x$.

\section{Problem Formulation}\label{sec: problem}
Consider an unknown nonlinear time-invariant system 
\begin{equation}\label{eqn: dynamics}
    \dot{x}(t) = h_1(x(t), u(t)), \quad 
    y(t) = h_2(x(t),u(t)),
\end{equation}
where $h_1: \mathcal{X}\times \mathcal{U}\rightarrow \mathcal{X} \subset\mathbb{R}^n$, { $h_2:\mathcal{X}\times \mathcal{U}\rightarrow \mathcal{Y}\subset \mathbb{R}^{n}$}, and  $x(t)\in  \mathcal{X} 
 \subset \mathbb{R}^n$, $u(t)\in\mathcal{U}\subset\mathbb{R}^m$, and $y(t)\in\mathcal{U}\subset\mathbb{R}^{n}$ are the state, input, and output vectors at time $t\in \mathbb{R}_{+}$ respectively. We assume  $ h_2(x(t),u(t))=x(t), \forall t\in \mathbb{R}_{+}$. The input of the system evolves as 
\begin{equation} \label{eqn: input}
    \dot{u}(t) = g(x(t), u(t)),
\end{equation}
allowing us to consider closed-loop identification with any time-invariant control input. 
{ We assume that $h_1, h_2,$ and $g$ are  twice differentiable.}
Stacking the state and input as ${z}(t){\triangleq} [x^T(t) \: \,  u^T(t)]^T{=}[y^T(t) \: \,  u^T(t)]^T,$
    we rewrite \eqref{eqn: dynamics}-\eqref{eqn: input} as
{\begin{equation} \label{eqn: autonomous}
\dot{z}(t)=\begin{bmatrix}
         h_1^T(x(t), u(t)) \; \;
         g^T(x(t), u(t)) 
    \end{bmatrix}^T \triangleq   f(z(t)).
\end{equation}}We assume that the nonlinear system \eqref{eqn: dynamics} is incrementally dissipative, with the notion defined as follows. 
\begin{definition} \label{def: qsr_dissipativity} The nonlinear system  (\ref{eqn: dynamics}) is said to be \textit{$QSR$-incrementally dissipative} or \textit{incrementally dissipative} in short, if for all output pairs $y_1(t), y_2(t)\in \mathcal{Y}$ and input pairs $u_1(t), u_2(t) \in \mathcal{U}$, for all $t\in \mathbb{R}_{+}$, we have
\begin{align} \label{eqn: incremental_dissipativity}
    \begin{bmatrix}
         \Delta y(t) \\
         \Delta u(t) 
    \end{bmatrix}^T   
       \begin{bmatrix}
         Q & S \\
         S^T & R
    \end{bmatrix}
    \begin{bmatrix}
         \Delta y(t) \\
         \Delta u(t) 
    \end{bmatrix} \geq 0,
\end{align}
where $\Delta y(t) = \delta(y_1(t),y_2(t))$ and $\Delta u(t) = \delta(u_1(t),u_2(t))$.
\end{definition}
For the remainder of the paper, we omit the dependence of all quantities on time $t$ for simplicity of notation. 

Incremental dissipativity, as defined above, extends the classical notion of $QSR$-dissipativity \cite{hill1976stability} to system trajectories rather than equilibrium points, and is particularly useful in analyzing far from equilibrium behavior where linear models fail, or oscillatory behavior where equilibrium is never achieved but system trajectories in non-equilibrium regimes remain  ``well-behaved''. We are interested in the problem of identifying a model for \eqref{eqn: dynamics} that preserves its dissipativity properties, since classical $QSR$-dissipativity (and its incremental version in \eqref{eqn: incremental_dissipativity}) can be used to guarantee a variety of useful input-output properties (or their incremental versions) through appropriate choices of the $Q,S,$ and $R$ matrices (see Remark below).

\begin{remark}\label{rem:qsr}
A wide variety of useful system properties can be guaranteed from $QSR$-dissipativity through appropriate choices of the dissipativity matrices, such as \cite{agarwal2019compositional}:
    \begin{enumerate}
        \item[(i)]
        $\mathcal{L}_2$ stability: $Q = -\frac{1}{\gamma} I, S = 0, R=\gamma I$, where $\gamma > 0$ is the $\mathcal{L}_2$ gain of the system;
        \item[(ii)] Passivity:
        $Q =0, S=\frac{1}{2}I, R=0$;
        \item[(iii)] Strict Passivity: $Q = -\epsilon I, S=\frac{1}{2}I, R=-\delta I$, where $\epsilon>0$ and $\delta>0$;
        \item[(iv)] Conicity: $Q = -I, S=cI, R=(r^2-c^2)I$, where $c\in \mathbb{R}$ and $r>0$.
        \item[(v)] {Sector-boundedness}: $Q=-I$, $S = (a+b)I$ and $R = -ab I$, where $a,b\in \mathbb{R}$.
    \end{enumerate}
    \end{remark}

Our main objective is to learn a neural dynamical system that closely approximates the behavior of the closed loop dynamics \eqref{eqn: autonomous} while preserving the incremental dissipativity of the unknown nonlinear system. Formally, the problem is formulated as identifying a \textit{neural dynamical system}
\begin{equation}\label{eqn: problem}
\dot{z}=\tilde{f}(z)
\end{equation}
satisfying \eqref{eqn: incremental_dissipativity} with 
 $\Delta z \triangleq \begin{bmatrix}
         \Delta y^T &
        \Delta u^T 
\end{bmatrix}^T$, and $\tilde{f}$ is a feed-forward fully-connected neural network (NN) with layers $\mathbf{L}_i, i\in\{1,2,\ldots,l\}$, whose mapping is defined as 
\begin{equation}\label{eqn: neural netowrk}
    \mathbf{L}_i: \quad z^i = \phi(v_i)  \quad \forall i\in \mathbb{Z}_{l}, \quad 
    \tilde{f}(z)= z^l,
\end{equation}
where $v_i=W_i z^{i-1}+b_i$, and $\phi:\mathbb{R} \to \mathbb{R}$ is a nonlinear \textit{activation function} that acts element-wise on its argument. The last layer $\mathbf{L}_l$ is termed the \textit{output layer} of the network. Our goal is to then learn the appropriate weights $W_i, i\in\mathbb{Z}_l$ and biases $b_i, i\in\mathbb{Z}_l$ that ensure that the neural dynamical system \eqref{eqn: problem} closely approximates the nonlinear system \eqref{eqn: autonomous}, while guaranteeing that it is incrementally dissipative in the sense of Definition \ref{def: qsr_dissipativity}. Note that identifying a neural dynamical system \eqref{eqn: problem} that closely approximates \eqref{eqn: autonomous} does not automatically guarantee that it is incrementally dissipative. 
\begin{remark}
    Intuitively, the phrase `closely approximates' means that the model behavior is very similar to the ground truth. Quantitatively, the model closely approximates the ground truth if 
    $\|y(t)-\hat{y}(t)\|_2 \leq \delta, \, \forall t\in \mathbb{R}_{+}$,
where $y(t)$ is the true output of the system at time $t$ and $\hat{y}(t)$ is the output of the model, and $\delta$ is some small user-defined tolerance. 
\end{remark}

\section{Learning Dissipative Neural Dynamics}\label{sec: learning}
\subsection{Identifying Neural Dynamics with No Constraints} \label{baseline}
Given $d$ system trajectories with $N$ data points each, denoted by $\{(\hat y_{ij},\hat u_{ij})\}, i\in\mathbb{Z}_d, j\in\mathbb{Z}_N$  on time interval $t \in [0,T]$, $T \in \mathbb{R}_{+}$ capturing the behavior of the nonlinear system \eqref{eqn: autonomous} in a region of interest, we create a training dataset formatted as $M$ collections $\{(y_j,u_j)^{(i)}\}, i \in \mathbb{Z}_M, j \in \mathbb{Z}_N$, where each collection comprises of consecutive data points sampled from any of the system trajectories starting at a randomly selected time point. A standard neural ODE training algorithm such as \cite{chen2018neural} can be used to identify a neural dynamical model comprised of a feed-forward fully-connected NN $\bar f$ with parameters $\bar \theta=(\bar W_i,\bar b_i),i \in \mathbb{Z}_l$, termed here as a \textit{baseline model}, and defined as
\begin{equation}  \label{eqn: Neural ODE}
    \begin{aligned}
    \dot z &= \bar f (z(t),\theta)  \quad t\in[0,T],
\end{aligned}
\end{equation}
that approximates the dynamical behavior of \eqref{eqn: autonomous}. 

As discussed earlier, there is no guarantee that the identified neural dynamical system \eqref{eqn: Neural ODE} is incrementally dissipative (Definition \ref{def: qsr_dissipativity}) even if the unknown nonlinear system \eqref{eqn: dynamics} is known to be dissipative. One approach to obtain a dissipative model is to constrain the NN parameters $\theta$ during training. However, typical neural ODE learning algorithms cannot directly handle constraints during training. Further, guaranteeing dissipativity properties such as \eqref{eqn: incremental_dissipativity} on the trained model requires imposing matrix inequality constraints on the training of neural ODE models; this is a complex problem for which no known algorithms exist. To address this issue, we propose an algorithm to perturb the parameters of the the baseline model post-training to guarantee incremental dissipativity, while retaining the fit of the learned model. 

\subsection{Dissipativity of Neural Dynamical Systems}
We first derive a matrix inequality condition on the NN weights that is sufficient to guarantee incremental dissipativity of the model. 
We  take advantage of a slope-restrictedness property on the activation function defined as follows. 
\begin{assumption}\label{assm: slope_restrictedness}
For the NN described in (\ref{eqn: neural netowrk}), the activation function $\phi$ is \textit{slope-restricted} in $[\alpha, \beta],$ where $\alpha < \beta$, that is, $\forall v_a, v_b\in \mathbb{R}^n$, we have element-wise
\begin{align} \label{ineq: slope-restrictedness elementwise}
    \alpha(v_b-v_a)\leq\phi(v_b)-\phi(v_a) \leq \beta(v_b-v_a),
\end{align}
As a result, we have 
\begin{equation} \label{ineq: slope}
\begin{bmatrix}
            v_b-v_a \\
         \phi(v_b)-\phi(v_a) 
         \end{bmatrix}^T
       \begin{bmatrix}
         pI & -mI \\
         -mI & I
        \end{bmatrix}
\begin{bmatrix}
          v_b-v_a\\
         \phi(v_b)-\phi(v_a)
       \end{bmatrix} \leq 0,
\end{equation}
where $p = \alpha\beta$ and $m=\frac{\alpha+\beta}{2}$ (see Appendix for derivation.)
\end{assumption}

Slope-restrictedness is satisfied by most widely-used activation functions. For example, for the ReLU, sigmoid, tanh, exponential linear functions, $\alpha=0$ and $ \beta=1$. For the leaky ReLU function, $\phi(x)=\max(ax,x)$, with $a>0$,  $\alpha=\min(a,1)$ and $\beta = \max(a,1)$ \cite[Proposition 2]{fazlyab2020safety}.

We can now derive the following condition from the slope-restrictedness of the activation function in Assumption \ref{assm: slope_restrictedness}. 

\begin{figure*}[t]
{\scriptsize
    \setcounter{equation}{9}
    \vspace{0.5em}
\begin{align} \label{ineq: ML} 
M_L = 
\begin{bmatrix}
        P_{11}+\lambda_1\lambda p W_1^TW_1 & -\lambda_1\lambda m W_1^T & 0 &... & P_{12}\\
        -\lambda_1\lambda m W_1 & \lambda_1\lambda I+\lambda_2\lambda p W_2^TW_2 & -\lambda_2\lambda m W_2^T &... & 0\\
        0 & -\lambda_2\lambda m W_2 & \lambda_2\lambda I+\lambda_3\lambda p W_3^TW_3 & ... & 0\\
        & & \vdots & &\\
        0 & ... & 0 & \lambda_{l-1}\lambda I + \lambda_l \lambda p W_l^TW_l & -\lambda_l \lambda m W_l^T\\
        P_{21} & 0 & ... & - \lambda_l \lambda m W_l & P_{22}+\lambda_l\lambda I
\end{bmatrix}  \geq 0 
\end{align}}
    \vspace{-1em}
        \hrulefill
\end{figure*}
\setcounter{equation}{10}

\begin{lemma} \label{lemma: input output quadratic}
For the NN \eqref{eqn: neural netowrk}, if there exist $\lambda_i \in \mathbb{R}_{+}, i \in \mathbb{Z}_l$ and $\lambda \in \mathbb{R}_{+}$ satisfying \eqref{ineq: ML}, with $P_{11}$, $P_{22}$ being symmetric matrices, and $P_{12}^T = P_{21}$, then 
\begin{align} \label{ineq: system A}
    \begin{bmatrix}
       \Delta z^0 \\
       \Delta z^l
\end{bmatrix}^T  
   \begin{bmatrix}
        P_{11} & P_{12}\\
        P_{21} & P_{22}
\end{bmatrix}
   \begin{bmatrix}
       \Delta z^0\\
    \Delta z^l
\end{bmatrix}\geq 0,
\end{align}
 where $\Delta z^0 = \delta(z^0_1,z^0_2)$ and $\Delta z^l = \delta(z^l_1,z^l_2)$, where $(z^0_1,z^l_1)$ and $(z^0_2,z^l_2)$ are  input-output pairs for the NN defined in (\ref{eqn: neural netowrk}).
\end{lemma}

Finally, we are ready to derive a sufficient condition for incremental dissipativity of the neural dynamics \eqref{eqn: problem}. 
\begin{theorem} \label{theorem: main theorem}
If there exist appropriate $(Q,S,R)$ such that \eqref{ineq: ML} holds, with  $P_{11} = \begin{bmatrix}
        Q & S \\
        S^T & R 
\end{bmatrix}$, $P_{12}=P_{21}=0$, $P_{22}<0$,
then the system \eqref{eqn: problem} is $QSR$-incrementally dissipative in the sense of Definition \ref{def: qsr_dissipativity}, that is, it satisfies \eqref{eqn: incremental_dissipativity}.

\begin{remark}
    Note that verification of dissipativity using \eqref{ineq: ML} typically involves choosing the $Q,S,$ and $R$ matrices as optimization parameters that satisfy \eqref{ineq: ML}, with the appropriate parametrization chosen based on the desired property (such as passivity or sector-boundedness) in Remark \ref{rem:qsr}. 
\end{remark}
\end{theorem}

\subsection{Algorithm to Learn Dissipative Neural Dynamics}
\begin{figure}[t]
      \centering
      \includegraphics[scale=0.26,trim=0cm 1cm 0.1cm 1.2cm]{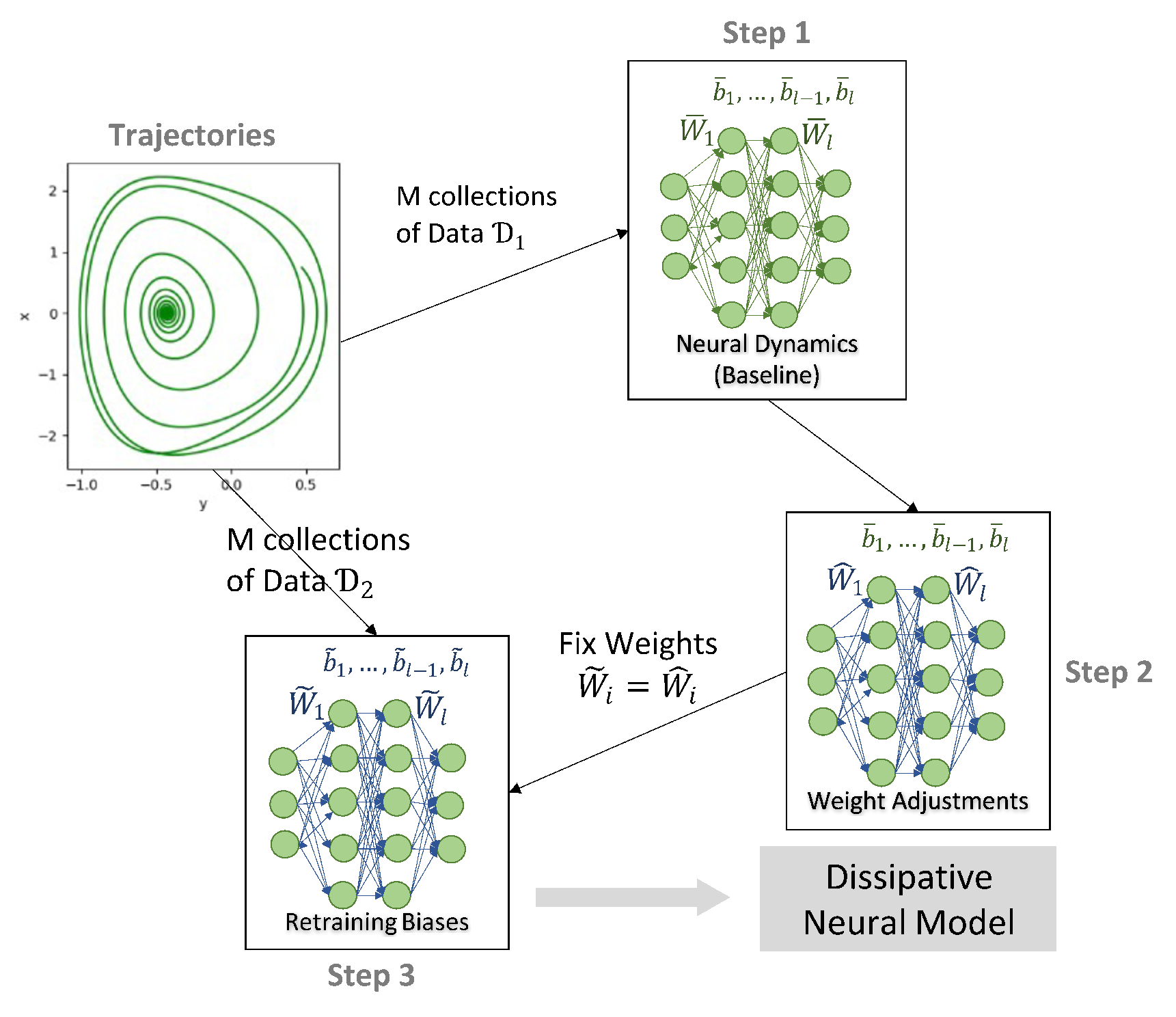}
      \caption{Approach to learn dissipative neural dynamics}
      \label{fig: schematic}
   \end{figure}

We now present the complete algorithm to learn a dissipative neural dynamical model that approximates the unknown nonlinear system \eqref{eqn: dynamics}, summarized in Fig. \ref{fig: schematic}. We first train the baseline model $\bar{f}$ with parameters $\bar \theta$ satisfying \eqref{eqn: Neural ODE} with no constraints as described in Section \ref{baseline}. 
Then, we perturb its weights to enforce incremental dissipativity. We would ideally like to minimize the dissipativity-enforcing weight perturbation, in order to maintain the closeness of the learned model to the behavior of the nonlinear system.  
We formulate the following optimization problem to realize this step:
\begin{equation} \label{weight perturbation}
\begin{aligned}
\hat W = \argmin_{W_1, W_2, ..., W_l} \quad & \sum\limits_{i=1}^l \|W_i-\bar{W}_i\|^2_2\\
\textrm{s.t.} \quad & M_L\geq 0, \quad 
  \lambda_i\geq 0 \quad i \in \mathbb{Z}_l, 
\end{aligned}
\end{equation}
where $M_L$ is defined in (\ref{ineq: ML}) and $P_{11}, P_{12}, P_{21}, P_{22}$ are chosen following Theorem \ref{theorem: main theorem}.
   
\begin{remark}\label{rem: independent}
   Note that enforcing dissipativity in our model only requires constraints on weights, and the dissipativity property still holds even if the biases are changed. This is due to the fact that the incremental dissipativity property in \eqref{eqn: incremental_dissipativity} is with respect to the difference in the inputs and the biases cancel out when we derive the sufficient condition for \eqref{ineq: system A} in Lemma \ref{lemma: input output quadratic} (see proof in  the Appendix).
\end{remark}

The last step is to further adjust the biases to compensate for any loss of fit to the original nonlinear system due to the weight perturbation. We re-sample the  trajectory data to avoid over-fitting by not using the same training data as in the first step. Then, we freeze the weights $\hat{W}_i$ and train only the biases using the new sampled data. The training yields biases $\tilde{b}_i, i \in \mathbb{Z}_l$. The final model $\tilde{f}$ has parameters $\tilde{W}_i=\hat{W}_i$, and $\tilde{b}_i, i \in \mathbb{Z}_l$. Algorithm \ref{alg: learning} summarizes this procedure.
\setlength{\textfloatsep}{10pt}%
\begin{algorithm} 
\caption{Dissipative Neural Dynamics Identification} \label{alg: learning}
    \textbf{Input} Two different data sets $\mathcal{D}_1$ and $\mathcal{D}_2$, each comprising of $M$ collections $\{(y,u)^{(i)}\}_{i=1}^M$ on fixed-length time intervals 
    \\
    \textbf{Output} Incrementally dissipative neural dynamical model \eqref{eqn: problem},  parameters $(\tilde W_i, \tilde b_i)$, $i{\in}\mathbb{Z}_l$ 
\begin{algorithmic}[1] 
\STATE Using  $\mathcal{D}_1$, train an unconstrained neural ODE model (baseline model) $\bar{f}$ with parameters $(\bar{W}_i,\bar{b}_i), i \in \mathbb{Z}_l$.
\STATE Solve problem (\ref{weight perturbation}) with $M_L$ defined as in (\ref{ineq: ML}) and $P_{11}, P_{12}, P_{21}, P_{22}$  chosen according to Theorem \ref{theorem: main theorem}. Obtain model $\hat{f}$ with  parameters $(\hat{W_i},\bar{b}_i), i \in \mathbb{Z}_l$.
\STATE Set weights $\tilde W_i \leftarrow \hat{W_i}$.
\STATE Using  $\mathcal{D}_2$, retrain only the biases in \eqref{eqn: Neural ODE} and obtain $\tilde{b}_i$. 
\RETURN NN parameters $(\tilde W_i, \tilde b_i), i \in \mathbb{Z}_l$.
\end{algorithmic}
\end{algorithm}

\section{Case Study}\label{sec: case}
We provide a numerical example on a second-order Duffing oscillator to illustrate the proposed learning approach.

\textit{Second-order Duffing Oscillator} \cite[Example 23]{verhoek2023convex}:  
The nonlinear dynamics of the Duffing oscillator is given by:
\begin{equation}\label{eqn: duffing}
\begin{aligned}
\dot{x}_1(t) &= x_2(t) \\
\dot{x}_2(t) &= -ax_2(t)-(b+cx_1^2(t))x_1(t)+u(t),
\end{aligned}
\end{equation}
where $x_1$ and $x_2$ are the states, $u$ is the control input, and $a,b$ and $c$ are parameters, chosen as  $a=1,b=1,c=1$. The system displays nonlinear oscillatory behavior, making neural dynamical models an attractive candidate to capture the dynamics. Further, the system is known to be incrementally dissipative, which is a property that we would like to preserve in the learned model. We implement Algorithm \ref{alg: learning} to learn a dissipative neural dynamical model in three steps.

\begin{figure*}[htbp]
\centering
\vspace{0.75em}
\begin{minipage}[t]{0.31\textwidth}
\centering
\includegraphics[scale=0.4,trim=0.5cm 0.5cm 0cm 0cm]{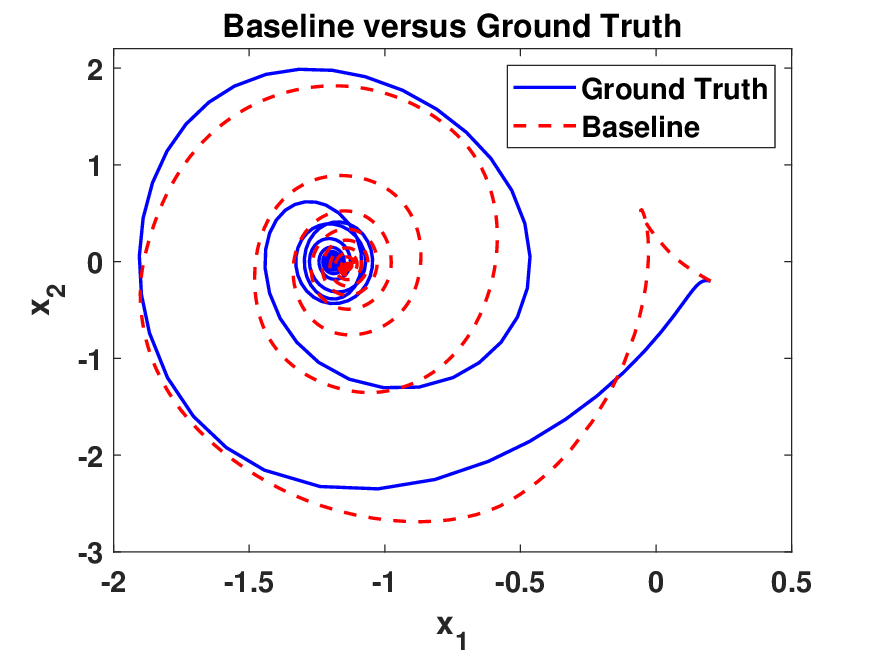}
\caption{Trajectories of the baseline (unconstrained) neural ODE model and the ground truth for a test input.}
\label{fig: baseline}
\end{minipage}
\hspace{0.01\textwidth}
\begin{minipage}[t]{0.31\textwidth}
\centering
\includegraphics[scale=0.4,trim=0.5cm 0.4cm 0cm 0cm]{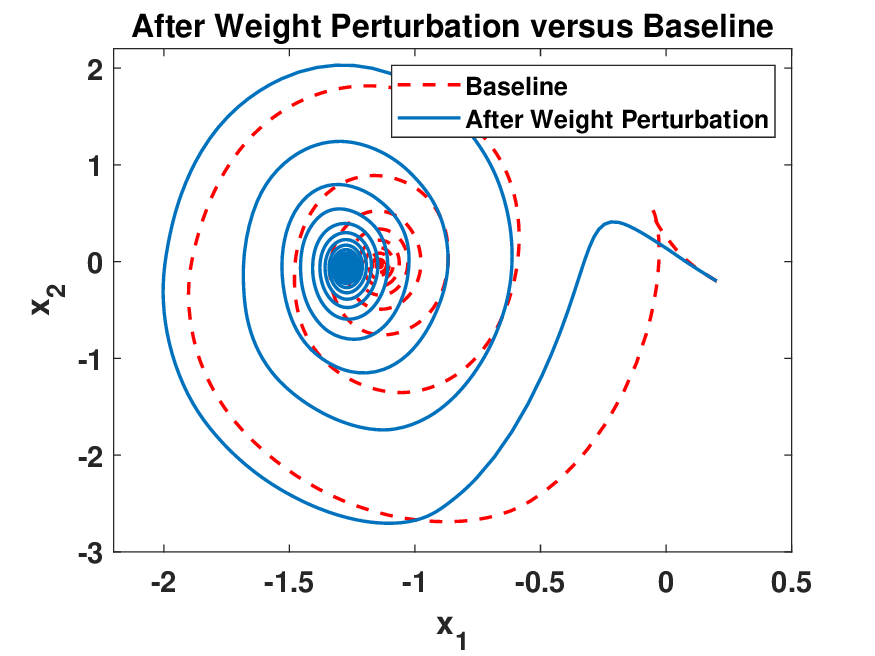}
\caption{Trajectories of baseline model and the dissipative model  after weight perturbation.}
\vspace{0.5em}
\label{fig: weight perturbation}
\end{minipage}
\hspace{0.01\textwidth}
\begin{minipage}[t]{0.31\textwidth}
\centering
      \includegraphics[scale=0.4,trim=0.5cm 0.4cm 0cm 0cm]{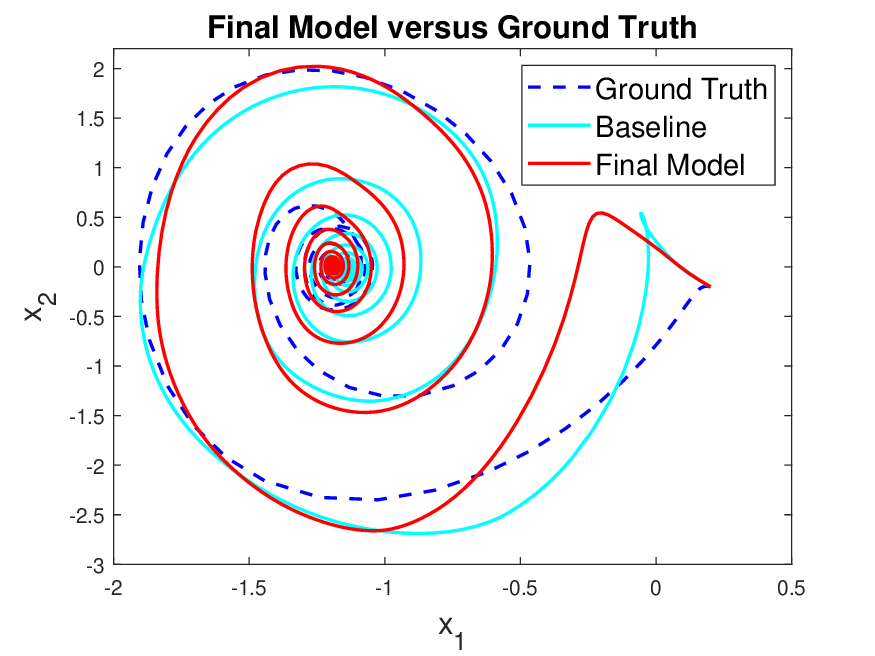}
      \caption{Trajectories of the dissipative model after bias adjustment, baseline model, and the ground truth.}
      \label{fig: final model}
      \end{minipage}
      \vspace{-1em}
\end{figure*}

\textit{Learning a baseline model:} We begin by learning an unconstrained baseline neural ODE model, using the algorithm in \cite{chen2018neural}. We pick inputs with $\dot{u}(t) = (0.6e^{-0.2t}cos(\pi t)-3\pi e^{-0.2t}sin(\pi t)) \mathbf{1}(t),$
where $\mathbf{1}(t)=1$ for $t\in \mathbb{R}_{+}$ and 0 otherwise. We pad the input with a dummy variable, set to  zero at all times, ensuring that the augmented input has the same dimension as the state vector. We use a feed-forward fully-connected NN with 1 hidden layer of 16 neurons. 
Note that the output layer of the NN does not have an activation function (that is, $\alpha=\beta=1$ for the the output layer). We simulate three trajectories with randomly assigned initial conditions and inputs starting from $0$, $0.1$ and $0.2$ respectively. For each trajectory, we obtain 10000 evenly distributed data points. Then, we form 100 data collections by randomly selecting  time intervals, each containing 6000 consecutive data points.  We add Gaussian noise $n\sim \mathbf{N}(0,0.01)$  to the states and input to emulate noisy sensor data often encountered in practice. Fig. \ref{fig: baseline} shows our baseline model, which closely approximates the ground truth. We test the dissipativity of the baseline model, and find no feasible $Q=-\epsilon I, R=-\delta I, \epsilon,\delta \geq 0$ satisfying \eqref{ineq: ML} even though the original system \eqref{eqn: duffing} is incrementally passive in the sense of Remark \ref{rem:qsr}-(iii). By relaxing the dissipativity condition, the best dissipativity indices we can obtain, in the sense of minimizing $\min(-\epsilon,0)+\min(-\delta,0)$, 
are $\epsilon=-0.0837,\delta=-0.1281<0$, indicating that we cannot establish  incremental passivity of the baseline model. This illustrates that dissipativity is not naturally inherited, even if the system dynamics is accurately approximated.

\textit{Weight perturbation to enforce dissipativity:}
Despite the nonlinear system \eqref{eqn: duffing} being incrementally dissipative, the baseline model fails to preserve this property. Therefore, in the second step, we solve the optimization problem described in \eqref{weight perturbation}  to obtain dissipative neural dynamical model.  Particularly, we impose the property of incremental passivity by setting $S = 0.5\mathbf{I}$, and choosing $Q=-\epsilon I, R=-\delta I, \epsilon,\delta \geq 0.$
We choose the negative definite matrix $P_{22}=-0.01\mathbf{I}$. Using YALMIP/Penlab \cite{lofberg2004yalmip}\cite{fiala2013penlab} to solve \eqref{weight perturbation}, we obtain a dissipative neural dynamical system with $R = -9.9168\times10^{-6} \mathbf{I}$, $Q = -9.9564 \times 10^{-6} \mathbf{I}$, $\lambda_1 = 10.5945$, and $\lambda_2 = 17.9737$. The 2-norm of the perturbation on the flattened weight variables (a 128-dimensional vector) is 1.4443. The dissipative model obtained after weight perturbation is tested on the same trajectory and compared with the baseline model in Fig. \ref{fig: weight perturbation}. We observe that we manage to impose dissipativity through just a small perturbation.

\textit{Bias Adjustment}:
Despite the perturbation being small, it may still drive the model away from the ground truth to some extent. This is because the neural ODE is nonlinear, and small parametric changes may still lead to non-negligible output deviations that accumulate when the ODE is integrated to obtain system trajectories. Therefore, in the last step, we freeze the weights (which were designed to guarantee dissipativity), and adjust only on biases to compensate for any loss of fit. Note that the biases can be trained independently while maintaining dissipativity guarantees (Remark \ref{rem: independent}). We collect training data in a similar manner as the first step, but pick starting points for generating the three trajectories using a different random seed. We purposely do this to avoid overfitting. After bias adjustment, we demonstrate that our final model closely matches the ground truth and retains closeness to the unconstrained baseline model (Fig. \ref{fig: final model}), while guaranteeing  incremental dissipativity.

\section{Conclusion and Future Work}\label{sec: future}
We presented an approach to learn neural dynamical models for nonlinear systems that preserve system dissipativity properties. Our approach involves first learning a baseline neural ODE model, followed by a minimal perturbation to enforce dissipativity, while retaining model fit. Future directions include compositional approaches for weight adjustments to decrease computational cost, and 
 utilizing our sufficient conditions for dissipativity to design controllers that satisfy closed-loop dissipativity specifications.

\section{Appendix}
We state the proofs of all results here. An important tool is the lossless S-procedure, stated below.
\begin{lemma}[S-Procedure] \label{S procedure}
For two matrices $F_0=F_0^T$ and $F_1=F_1^T$, if there exists $\lambda \in \mathbb{R}^+$ such that $F_0\geq \lambda F_1$, then $z^TF_1z\geq0,\forall z$ implies $z^TF_0z\geq 0, \forall z$. Additionally, if there exists a vector $z_0$ such that $z_0^T F_1 z_0>0$, 
then the converse holds, that is, $z^TF_0z\geq 0, \forall z$ implies $z^TF_1z\geq0,\forall z$.
\end{lemma}
\begin{figure*}[t]
{\scriptsize
    \setcounter{equation}{16}
    \vspace{0.5em}
\begin{align}\label{eqn: ST}
   S_T= \begin{bmatrix}
    \lambda_1 p W_1^TW_1 & -\lambda_1 m W_1^T & 0 &... & 0\\
        -\lambda_1 m W_1 & \lambda_1 I+\lambda_2 p W_2^TW_2 & -\lambda_2 m W_2^T &... & 0\\
        0 & -\lambda_2 m W_2 & \lambda_2 I+\lambda_3 p W_3^TW_3 & ... & 0\\
        & & \vdots & &\\
        0 & ... & 0 & \lambda_{l-1} I + \lambda_l p W_l^TW_l & -\lambda_l  m W_l^T\\
        0 & 0 & ... & - \lambda_l  m W_l & \lambda_l I\end{bmatrix}
\end{align}
        \hrulefill
        }
\end{figure*}
\setcounter{equation}{13}

\textit{{Proof of Lemma \ref{lemma: input output quadratic}}}:
For each layer $\mathbf{L}_i$, $i\in\mathbb{Z}_l$, define $\Delta z^i = \delta(z_a^i,z_b^i)$, where $z_a^i$ and $z_b^i$, such that $z_a^i\neq z_b^i$, are two different inputs to the NN layer $\mathbf{L}_i$. Similarly, define $\Delta v_i=\delta(v_a^i,v_b^i)$, where $v_a^i$ and $v_b^i$ are the linear transformations of $x_a^{i-1}$ and $x_b^{i-1}$, defined as $v_a^i=W_ix_a^{i-1}+b_i$ and $v_b^i=W_ix_b^{i-1}+b_i$. From \eqref{ineq: slope}, for any layer $\mathbf{L}_i$, $i\in\mathbb{Z}_l$,  
\begin{align} \label{ineq: slope in proof}
{\small\begin{bmatrix}
          \Delta v^{i}\\
         \Delta \phi(v^{i})
    \end{bmatrix}^T   
       \begin{bmatrix}
         pI & -mI \\
         -mI & I
    \end{bmatrix}
    \begin{bmatrix}
          \Delta v^{i}\\
         \Delta \phi(v^{i})
    \end{bmatrix}  \leq 0}
\end{align}
Notice that 
$
    \Delta v^{i}=(W_iz_b^{i-1}+b_i)-(W_iz_a^{i-1}+b_i) = W_i\Delta z^{i-1}
$
and $\Delta \phi(v^{i}) = \Delta z^{i}$. We can rewrite 
\begin{equation*}
 {\small   \begin{bmatrix}
          \Delta v^{i}\\
         \Delta \phi(v^{i})
    \end{bmatrix} =
    \begin{bmatrix}
          W_i & 0\\
         0 & I
    \end{bmatrix}
    \begin{bmatrix}
          \Delta z^{i-1}\\
         \Delta z^{i}
    \end{bmatrix}.}
\end{equation*}
Substituting in \eqref{ineq: slope in proof}, we have
for any $\lambda_i\in\mathcal{R_+}
$\begin{equation} \label{ineq: one layer}
{\small    -\begin{bmatrix}
          \Delta z^{i-1}\\
         \Delta z^{i}
    \end{bmatrix}^T
    \begin{bmatrix}
          \lambda_i pW_i^TW_i & -\lambda_imW_i^T\\
         -\lambda_i mW_i & \lambda_iI
    \end{bmatrix}
    \begin{bmatrix}
          \Delta z^{i-1}\\
         \Delta z^{i}
    \end{bmatrix} \geq 0 }
\end{equation}
Stacking the inequalities for all layers in a diagonal manner, 
\begin{equation} \label{ineq: stack}
{\small  \begin{bmatrix}
          (\Delta z^{0})^T 
          \dots 
         (\Delta z^{l})^T
    \end{bmatrix} (-S_T)
     \begin{bmatrix}
          (\Delta z^{0})^T 
          \dots
         (\Delta z^{l})^T
    \end{bmatrix}^T\geq  0,}
\end{equation}
where $S_T$ is defined in \eqref{eqn: ST}. Using the S-procedure  in Lemma \ref{S procedure}, under mild conditions (discussed shortly), if there exists a non-negative $\lambda\in\mathbb{R}$ such that 
\setcounter{equation}{17}
\begin{equation} \label{ineq: S procedure application}
{\scriptsize   P_L{\triangleq}\begin{bmatrix}
    P_{11} & 0 &... & 0 & P_{12}\\
    0 &&...&& 0\\
    & &\vdots&& \\
       0 &&...&& 0\\
    P_{21} & 0 & ... & 0 & P_{22} \end{bmatrix}\geq-\lambda  S_T,}
\end{equation}
\noindent then  $  \begin{bmatrix}
          (\Delta z^{0})^T
          \dots
         (\Delta z^{l})^T
    \end{bmatrix}
   P
     \begin{bmatrix}
          (\Delta z^{0})^T
          \dots
         (\Delta z^{l})^T
    \end{bmatrix}^T{\geq}0,$
implying \eqref{ineq: system A}.
Notice that the  condition in \eqref{ineq: S procedure application} is exactly $M_L$. From Lemma \ref{S procedure}, we require a mild condition, namely the existence of $\Delta z^i$, $i\in\mathbb{Z}_l$, such that \eqref{ineq: stack}
 holds strictly. As $\alpha < \beta$ and the inputs are different, there exists some $\Delta z^{0}, \Delta z^{1}$ such that \eqref{ineq: one layer} is strict. Then with any $\Delta z^{i}$, $i\in \{2,...,l\}$ , \eqref{ineq: stack} holds strictly.

\textit{Proof of Theorem \ref{theorem: main theorem}}:
    With $P_{11} = \begin{bmatrix}
        Q & S \\
        S^T & R 
\end{bmatrix}$, $P_{12}=P_{21}=0$, and $P_{22}<0$, we can write \eqref{ineq: system A} as
\begin{equation}\label{eqn: ineq_thm1}
 {\small   (\Delta z^{0})^T \begin{bmatrix}
        Q & S \\
        S^T & R 
\end{bmatrix} \Delta z^{0} + (\Delta z^{l})^T P_{22} \Delta z^{l} \geq 0.}
\end{equation}
Note that $P_{22}$ is negative definite, which means $(\Delta z^{l})^T P_{22} \Delta z^{l} < 0$. Therefore, the first term in \eqref{eqn: ineq_thm1} is larger than 0. The conclusion directly follows from the fact that $\Delta z^{0} = \left[\Delta y^T(t), \Delta u^T(t)\right]^T$.

\textit{Proof of Relation in  Assumption \ref{assm: slope_restrictedness}:}
Inequality \eqref{ineq: slope-restrictedness elementwise} can be equivalently written as
$
    \left[(\phi(v_b)-\phi(v_a)) -\alpha(v_b-v_a)\right]_i \times
    \left[(\phi(v_b)-\phi(v_a))-\beta(v_b-v_a)\right]_i\leq 0, \quad \forall i\in \mathbb{R}^n,
$
where subscript $i$ means the $i^{th}$ element of the vector.
Summing all the inequalities for $i\in \mathbb{R}^n$, we have 
{\small\begin{equation*}
    \left[(\phi(v_b){-}\phi(v_a)){-}\alpha(v_b-v_a)\right]^T
    \left[(\phi(v_b){-}\phi(v_a)){-}\beta(v_b-v_a)\right]{\leq}0.
\end{equation*}}
\noindent In other words, we have the following quadratic inequality $(\phi(v_b){-}\phi(v_a))^T(\phi(v_b){-}\phi(v_a)){-}\frac{\alpha+\beta}{2}(\phi(v_b){-}\phi(v_a))^T(v_b-v_a)
{-}\frac{\alpha+\beta}{2}(v_b{-}v_a)^T(\phi(v_b){-}\phi(v_a)){+}\alpha\beta(v_b{-}v_a)^T(v_b{-}v_a)\leq 0$, which can be rewritten as \eqref{ineq: slope} with $p{=}\alpha \beta$ and $m{=}\frac{\alpha+\beta}{2}$.

\balance

\bibliographystyle{IEEEtran}
\bibliography{ref}

\end{document}